\documentclass{article}

\usepackage{arxiv}
\usepackage[utf8]{inputenc} % allow utf-8 input
\usepackage[T1]{fontenc}    % use 8-bit T1 fonts
\usepackage{hyperref}       % hyperlinks
\usepackage{url}            % simple URL typesetting
\usepackage{booktabs}       % professional-quality tables
\usepackage{amsfonts}       % blackboard math symbols
\usepackage{nicefrac}       % compact symbols for 1/2, etc.
\usepackage{longtable}
\usepackage{array}
\usepackage{microtype}      % microtypography
\usepackage{lipsum}		% Can be removed after putting your text content
\usepackage{graphicx}
\usepackage{verbatim}
\usepackage{doi}
\usepackage{authblk}
\usepackage{amsmath}
\usepackage{hyperref}
\hypersetup{
    colorlinks=false,  % 禁用链接颜色
    pdfborder={0 0 0}, % 禁用链接边框
    hidelinks          % 隐藏链接
}

\title{Animation Needs Attention: A Holistic Approach to Slides Animation Comprehension with Visual-Language Models}

\date{June 30, 2025}	% Here you can change the date presented in the paper title
\author[1]{Yifan Jiang}
\author[1]{Yibo Xue}
\author[1]{Yukun Kang}
\author[1]{Pin Zheng}
\author[1]{Jian Peng}
\author[2]{Feiran Wu}
\author[1]{Changliang Xu}

\affil[1]{Hangzhou Institute for Advanced Study, University of Chinese Academy of Sciences, Hangzhou 310024, P. R. China\thanks{Corresponding author: Changliang Xu (Email: xuchangliang@ucas.ac.cn; ORCID: https://orcid.org/0009-0000-2073-3429).}}
\affil[2]{Alibaba Group, Hangzhou, 311121, P. R. China}

\renewcommand{\shorttitle}{Animation Needs Attention}

\hypersetup{
pdftitle={ANIMATION NEEDS ATTENTION: A HOLISTIC APPROACH TO
SLIDES ANIMATION COMPREHENSION WITH VISUAL-LANGUAGE
MODELS},
pdfsubject={cs.AI},
pdfauthor={Yifan Jiang, Yibo Xue},
pdfkeywords={Slide, Animation,Multimodal Data Synthesis,Vision-Language Model,LoRA},
}

\begin{document}
\maketitle

\begin{abstract}
	Slide animations—such as fade-in, fly-in, and wipe—are essential for audience engagement, efficient information delivery, and vivid visual expression. Yet most AI-driven slide-generation tools still lack native animation support, and existing vision–language models (VLMs) underperform on animation tasks because no public datasets exist, and their temporal-reasoning ability remains limited. To close this gap, we release the first public dataset for slide-animation modelling: 12,000 triplets of natural-language descriptions, animation JSON files, and rendered videos that collectively cover every built-in PowerPoint effect. Using this resource, we fine-tune Qwen-2.5-VL-7B with Low-Rank Adaptation (LoRA) and achieve consistent gains over GPT-4.1 and Gemini-2.5-Pro on BLEU-4, ROUGE-L, SPICE, and our Coverage–Order–Detail Assessment (CODA) metric, which jointly gauges action coverage, temporal order, and detail fidelity. On a manually created slides test set, the LoRA model lifts BLEU-4 by $\approx${ 60} \% and ROUGE-L by $\approx${ 30} \%, while posting double-digit improvements in CODA-detail—evidence that low-rank adaptation confers reliable temporal reasoning and generalization beyond synthetic data. Overall, our dataset, LoRA-enhanced model, and CODA metric provide a rigorous benchmark and a preparation for future research on VLM-based dynamic slide generation.
\end{abstract}

% keywords can be removed
\keywords{Slide \and Animation \and Multimodal Data Synthesis \and Vision-Language Model\and LoRA (Low-Rank Adaptation)}

\clearpage

\section{Introduction}
Slides are ubiquitous in education, business, and scientific communication, prized for their clarity and flexibility. Animation effects—such as fade-in, fly-in, and wipe—add temporal dynamics to otherwise static content by precisely controlling the timing, direction, and style of each transition. These effects sharpen information delivery, capture audience attention, and enrich visual expression. Yet, despite the rise of AI-driven agent tools, most slide-generation systems still omit animation, limiting their utility for dynamic presentations. This study tackles that gap by introducing methods for the automatic understanding and generation of slide animations.

Animated slides differ fundamentally from plain text or static images: they comprise multi-frame sequences with strict temporal dependencies, demanding sophisticated multimodal reasoning and sequence modeling. Current vision–language models (VLMs) struggle for two main reasons: (1) no public datasets exist that target slide animations, hindering systematic training and evaluation; and (2) existing models have only limited ability to capture fine-grained motion cues and uphold temporal order.

Multimodal understanding has advanced rapidly in recent years. Datasets for document visual question answering—such as DocVQA\cite{mathew2021docvqa}, InfographicVQA\cite{mathew2022infographicvqa}, and SlideVQA\cite{tanaka2023slidevqa}—have spurred work on static documents and slides, whereas large-scale video-caption corpora (e.g., MSR-VTT\cite{xu2016msr} and ActivityNet Captions\cite{krishna2017dense}) have accelerated research on dynamic scene description. Yet methods aimed at text-heavy images (e.g., TextVQA\cite{singh2019towards}) still fail to model temporal dynamics, and open-domain video captioners seldom achieve the precision required for slide’s structured layouts and parameterized animations.

To close this gap, we release a synthetic dataset of 12,000 triplets—natural-language animation descriptions, animation JSON files, and rendered videos—explicitly created for training and evaluating VLMs on slide-animation understanding. On this resource we fine-tune Qwen 2.5-VL-7B\cite{bai2025qwen2} with Low-Rank Adaptation (LoRA), which adds only a small set of trainable weights yet markedly improves the model’s ability to capture fine-grained motion cues and maintain temporal order—particularly valuable in low-resource settings\cite{hu2022lora}. Finally, we propose \textbf{Coverage–Order–Detail Assessment (CODA)}, an LLM-based metric that scores action coverage, temporal order, and detail fidelity, offering a comprehensive evaluation of VLM performance on slide-animation recognition.

PowerPoint serves as a practical and widely adopted platform for modeling slide animations due to its rich built-in effects, structured layout system, and native support for timed transitions—all essential for capturing the temporal and visual characteristics of real-world presentations. We develop an end-to-end synthesis pipeline that combines large language models, the python-pptx library\cite{python-pptx}, and Visual Basic for Applications (VBA) to produce the first public dataset of 12,000 text–JSON–video triplets spanning a wide range of PowerPoint animation effects and layouts. Building on this resource, we fine-tune Qwen-2.5-VL-7B with a LoRA-based strategy that integrates multimodal inputs, markedly strengthening the model’s slide-animation understanding. We also introduce Coverage–Order–Detail Assessment (CODA) and show that the LoRA model consistently outperforms baselines on BLEU-4\cite{papineni2002bleu}, ROUGE\cite{lin2004rouge}, SPICE\cite{anderson2016spice}, and all CODA sub-scores, while generalizing well to manually created slides from diverse domains.\footnote{Dataset and code will be made available at https://github.com/pampas-lab upon acceptance of the manuscript.}

The remainder of the paper is structured as follows. \textbf{Section 2} surveys related work and its limitations for slide-animation recognition. \textbf{Section 3} describes the dataset construction pipeline and accompanying statistics. \textbf{Section 4} details our LoRA fine-tuning method. \textbf{Section 5} presents the experimental setup, CODA metric design, ablation studies, and comparative results. \textbf{Section 6} concludes the paper and discusses future directions.

\section{Related Work}

\subsection{Visual Language Models and Document Understanding}
The rapid advancement of Large Language Models (LLMs) has catalyzed significant progress in Visual Language Models (VLMs), enabling sophisticated cross-modal understanding capabilities. CLIP\cite{hessel2021clipscore} established foundational work by learning joint representations of images and text through contrastive learning, while BLIP-2\cite{li2023blip} introduced the Querying Transformer architecture to efficiently bridge visual encoders with language models. Instruction-following VLMs such as LLaVA\cite{elgendy2024geollava} and InstructBLIP\cite{instructblip} have demonstrated enhanced capabilities in following complex visual instructions, while Flamingo\cite{alayrac2022flamingo} has shown remarkable few-shot learning abilities. The Qwen 2.5-VL-7B model\cite{bai2025qwen2} adopted in this paper is a representative open-source model stemming from this technological trend. It possesses outstanding image-text understanding and conversational abilities, providing a solid foundation for our targeted fine-tuning.

In the domain of document understanding, Visual Question Answering (VQA) has emerged as a mature research area with specialized datasets including DocVQA\cite{mathew2021docvqa} for scanned documents, InfographicVQA\cite{mathew2022infographicvqa} for complex infographics, and TextVQA\cite{singh2019towards} for dense text understanding. The LayoutLM series\cite{xu2020layoutlm,xu2020layoutlmv2,xu2021layoutxlm,huang2022layoutlmv3} has significantly advanced structured document understanding by comprehending spatial relationships and textual content simultaneously, while UDOP\cite{tang2023unifying} has further unified document understanding through integrated vision, text, and layout processing. Most relevant to our work is SlideVQA\cite{tanaka2023slidevqa}, which targets static PowerPoint slides, requiring understanding of layout, content, and logical relationships within individual slides.

However, existing approaches are fundamentally limited to static documents, focusing primarily on spatial layout understanding and OCR capabilities. They lack the ability to process temporal information inherent in PowerPoint animations, including element entrance sequences, motion trajectories, and transition effects—a critical gap our work addresses.

\subsection{Video Understanding and Temporal Modeling}
PowerPoint animations can be conceptualized as specialized short videos, making video captioning research highly relevant. Large-scale datasets such as MSR-VTT\cite{xu2016msr} and ActivityNet Captions\cite{krishna2017dense} have driven progress in dynamic scene understanding, while complex datasets like YouCookII\cite{zhou2018towards} have advanced step-by-step temporal decomposition in procedural videos. In recent years, video-language models have demonstrated impressive capabilities: Video-ChatGPT\cite{maaz2023video} and VideoLLaMA\cite{zhang2023video} have shown strong performance in video understanding and dialogue, while X-CLIP\cite{ma2022x} has advanced video-text retrieval through cross-modal learning.

Despite their temporal modeling strengths, existing video description models face critical limitations when applied to slide animations: (1)\textbf{Domain Mismatch}—models excel at describing natural scene actions (e.g., "a person is running") but lack specialized vocabulary for structured animation effects (e.g., "a text box flies in from the left"); (2) \textbf{Text Content Neglect}—traditional methods inadequately handle the extensive textual content central to slide-based information delivery. These limitations have motivated refined evaluation methodologies like VCapsBench\cite{zhang2025vcapsbench}, which introduced fine-grained assessment across 21 dimensions, indicating the need for specialized frameworks tailored to structured video forms like slide animations.

\subsection{Efficient Fine-tuning and Evaluation Methodologies}
Parameter-Efficient Fine-Tuning (PEFT) techniques address the computational challenges of domain adaptation without full fine-tuning. LoRA\cite{hu2022lora} exemplifies this approach by introducing trainable low-rank matrices while freezing pre-trained parameters, achieving effective adaptation even with limited data\cite{gao2023llama}. Variants such as AdaLoRA\cite{zhang2023adalora} have improved adaptive rank allocation, while QLoRA\cite{dettmers2023qlora} has enabled efficient fine-tuning through quantization. Alternative PEFT methods including Prefix-tuning\cite{li2021prefix} and P-tuning\cite{liu2021p} have also demonstrated effectiveness across various tasks.

For evaluation, traditional metrics including BLEU-4\cite{papineni2002bleu}, ROUGE\cite{lin2004rouge}, SPICE\cite{anderson2016spice} are widely adopted for video description tasks\cite{INACIO2023100488}. However, these n-gram and scene graph-based metrics inadequately capture semantic dimensions crucial for slide animations, such as action coverage, temporal order, and detail fidelity. Following recent trends in LLM-based automated evaluation, we propose that LLM evaluators can more accurately assess description quality, leading to our development of CODA to complement traditional metrics with nuanced evaluation capabilities.

\clearpage

\section{Dataset Construction and Statistical Analysis}

To meet the training and evaluation needs of vision-language models (VLMs) on slide animation recognition, we employ an automated synthesis approach to build a large-scale dataset of 12,000 animation samples. In this section, we systematically present the design rationale of our synthesis framework, detail the data generation pipeline—comprising static slide creation followed by animation description and video rendering—and provide statistical analyses to demonstrate the dataset’s diversity and applicability.

\subsection{Synthesis Framework Design}
We designed a modular synthesis framework to encompass diverse slide animation types, layouts, and temporal characteristics. By combining automated generation with fine-grained control, the framework ensures sample diversity, consistency, and scalability. The process begins with static slide creation, followed by successive addition of animation effects and temporal configurations, and concludes with exporting videos alongside corresponding JSON files and natural language descriptions.

The framework integrates \texttt{python-pptx} for precise static element layout and uses pywin32 to invoke custom VBA scripts for JSON parsing and Microsoft PowerPoint manipulation (see Section 3.2.2). To generate coherent, accurate descriptions, we leverage the newly released GPT-4.1, which excels at instruction following and multimodal long-context understanding. This choice enables advanced prompt engineering, ensuring both efficient and precise content generation and animation description.

\subsection{Data Generation Pipeline}
\begin{figure}
	\centering
	\includegraphics[width=0.9\textwidth]{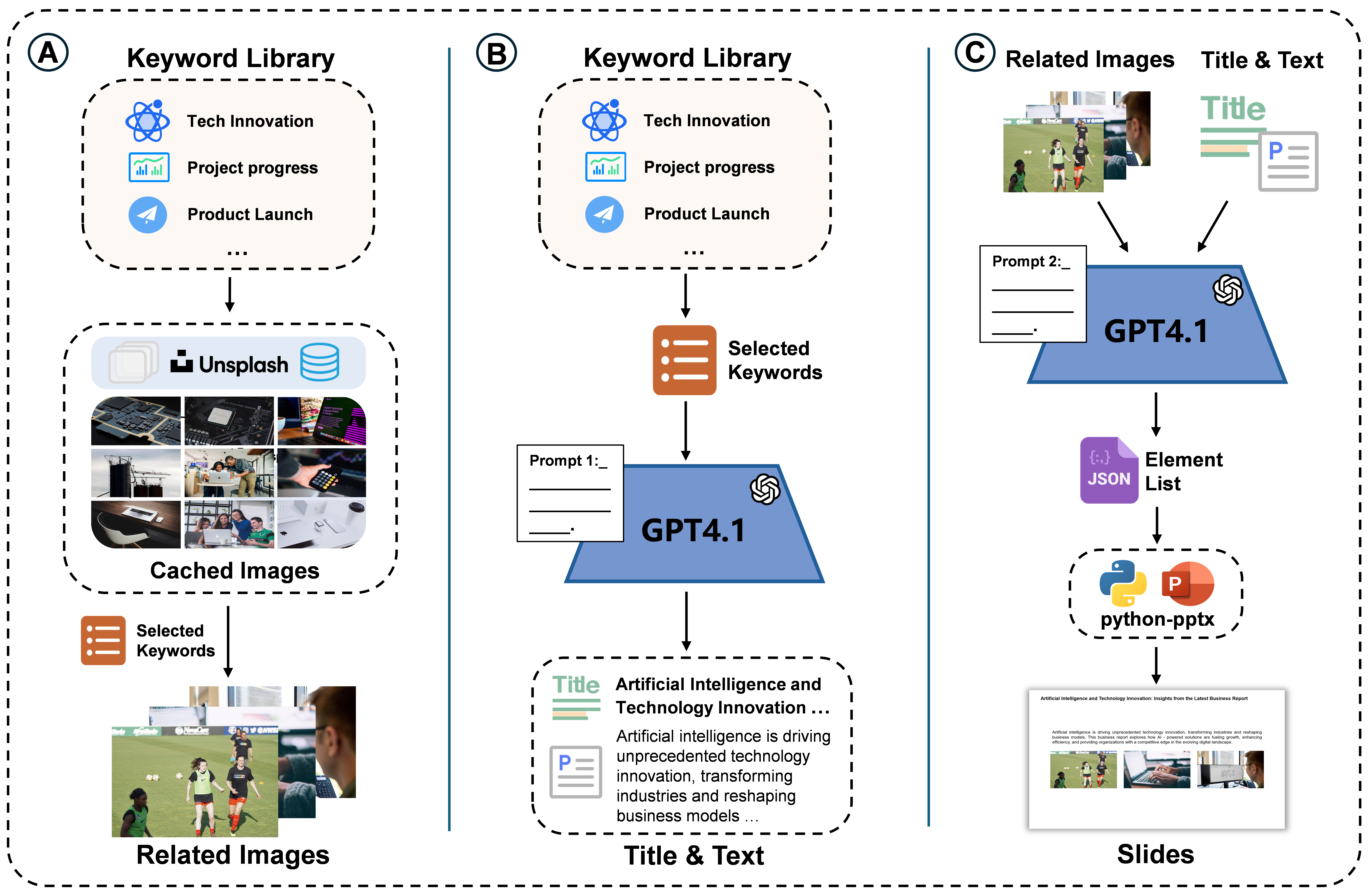}
	\caption{ Phase 1 pipeline for static slide synthesis: \textbf{(a) image pool construction, (b) slide text generation, and (c) slide rendering.}}
	\label{fig:fig1}
\end{figure}

Our pipeline produces 300 bilingual static slides and expands each into 40 animation variants, yielding 12,000 text–JSON–video triplets. It operates in two consecutive phases, illustrated in \textbf{Figure \ref{fig:fig1} (Phase 1)} and \textbf{Figure 2 (Phase 2)}.

\paragraph{Phase 1: Static Slide Synthesis}
We first construct a multilingual image pool by querying Unsplash with ten English–Chinese keyword pairs (40 images per keyword). 
For each slide, an LLM samples 1–4 keywords, drafts a title and body paragraph, and combines them with 1–4 randomly selected images to form an \texttt{element\_list}. The LLM then outputs a JSON layout specification—element name, x/y coordinates, width, and height—which python-pptx uses to render 300 static slides (150 English, 150 Chinese; layout-prompt details in \textbf{Appendix: Prompts}).

\begin{figure}
	\centering
	\includegraphics[width=0.9\textwidth]{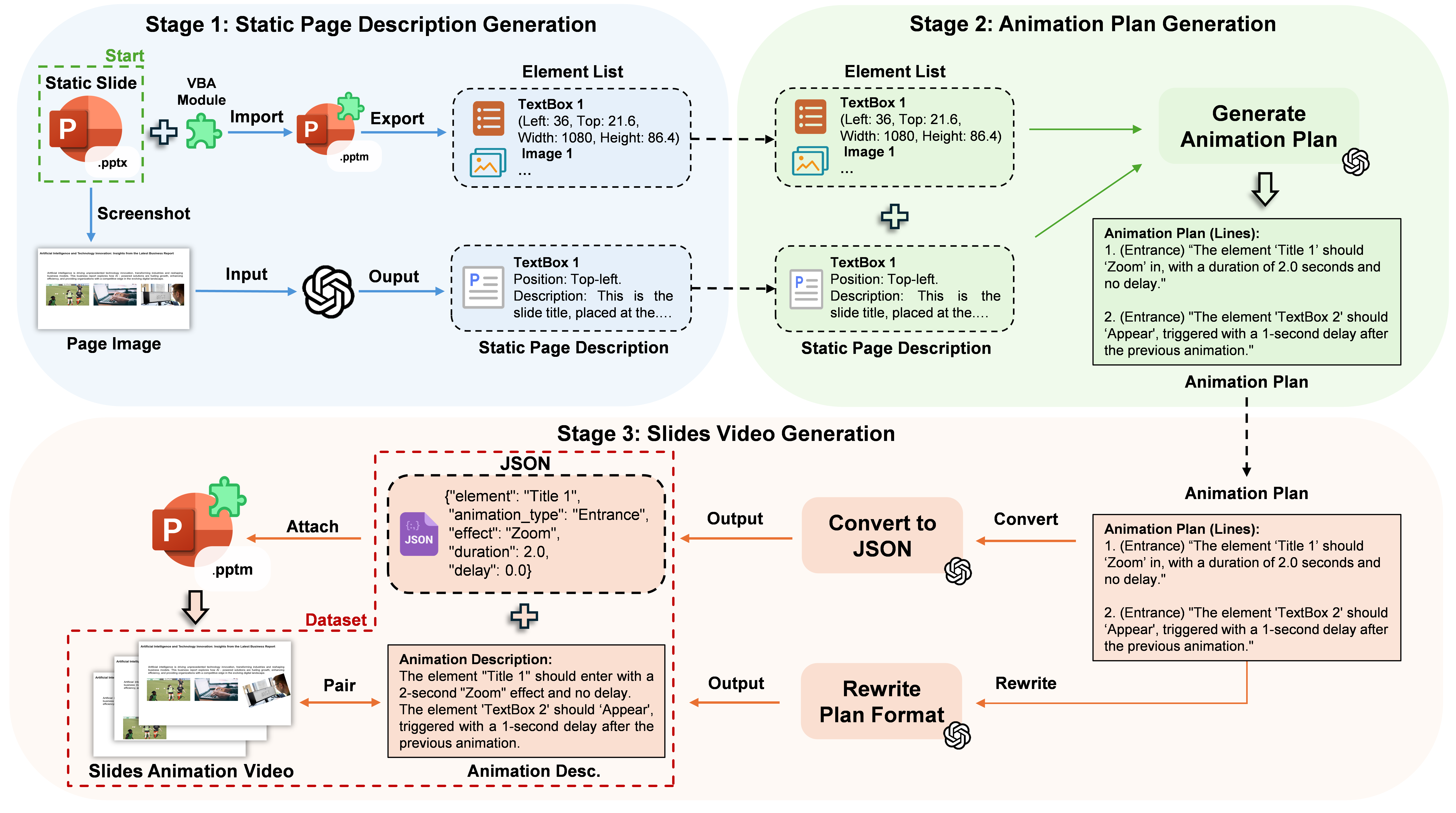}
	\caption{ Phase 2 pipeline with \textbf{Stage 1}: static page description generation, \textbf{Stage 2}: animation plan generation, and Stage 3: slides video generation.}
	\label{fig:fig2}
\end{figure}

\paragraph{Phase 2: Animation Description and Video Synthesis}
For each static slide we create \textbf{40 animation schemes}, producing \textbf{12,000 videos} that span every built-in PowerPoint effect except custom motion paths. Character-by-character text animations are excluded to keep runtimes reasonable; the final inventory covers \textbf{42 entrance–exit pairs} and \textbf{10 emphasis effects} (see \textbf{Appendix: Effect Table}). The pipeline comprises the three stages shown in \textbf{Figure \ref{fig:fig2}}.

\textbf{Static page description generation}: A screenshot and element list are extracted via VBA, then fed to GPT-4.1, which returns a detailed description of the static page.

\textbf{Animation plan generation}: The description is converted into a numbered, line-separated list of actions—index, animation type, element name, effect, duration, delay, and repeat count (e.g., “1. (Entrance) element ‘Title’ fades in over 1.5 s, 0 s delay, repeat 1”).

\textbf{Slides video generation}: The action list is paraphrased into a concise natural-language narrative for training input and converted into a structured animation JSON, which executed through PowerPoint’s COM interface by Five VBA modules: (1) element-list export, (2) JSON Parsing, (3) animation binding, (4) parameter setting, and (5) video export—ensuring smooth playback and strict temporal order.

\textbf{Table \ref{tab:table1}} presents a dataset sample: the natural-language animation description, the LLM-generated animation JSON applied via VBA, and the corresponding sequence of video frames.

\begin{table}[ht]
    \centering
    \renewcommand{\arraystretch}{1.8}  % Increase row height for better centering
    \begin{tabular}{@{}p{5cm} p{5cm} p{5cm}@{}}
    \toprule
    Natural Slide Animation Description & JSON & Video Frames      \\ \midrule
    \multicolumn{3}{c}{\includegraphics[width=0.95\textwidth]{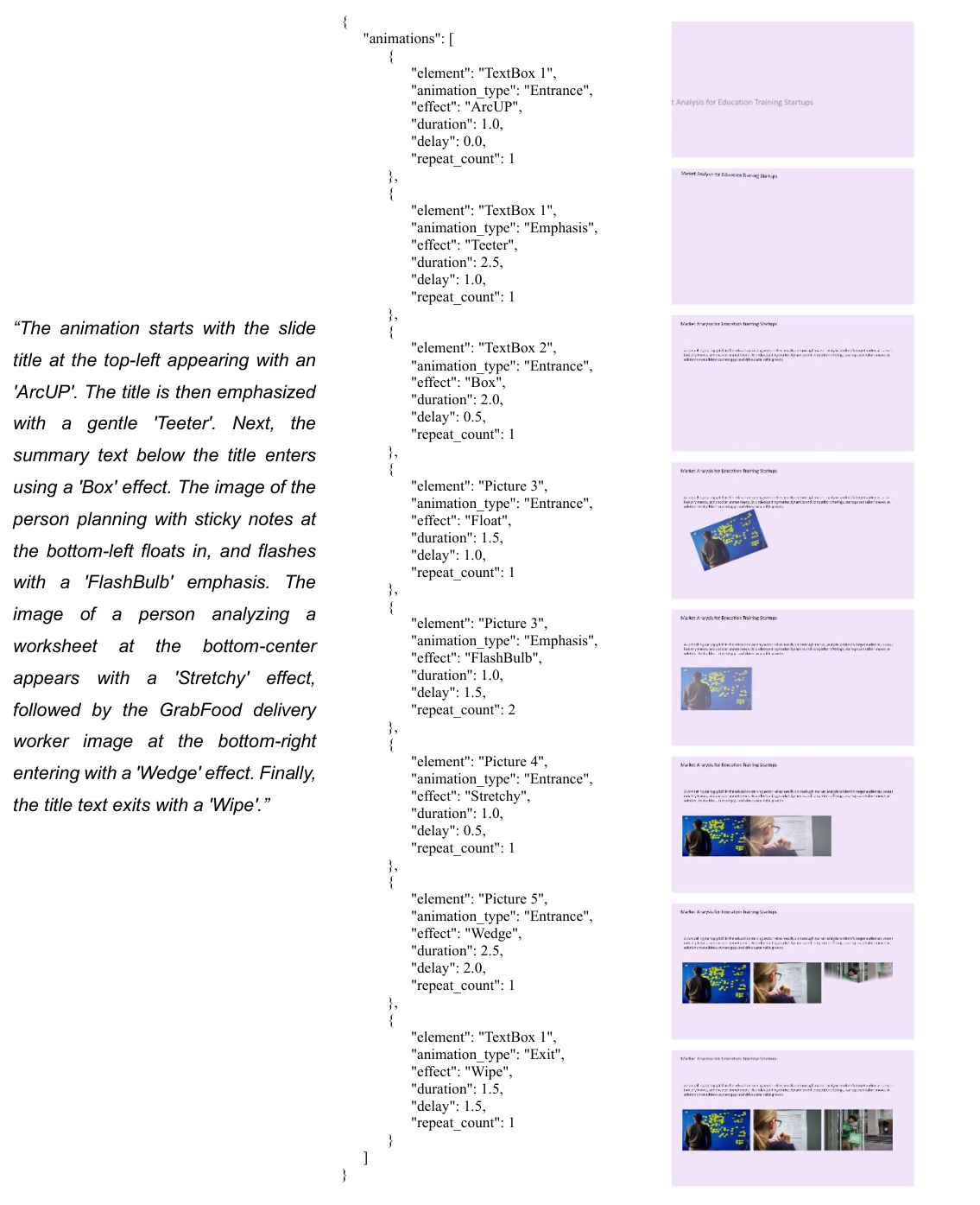}} \\ \bottomrule
    \end{tabular}
    \caption{Data sample of synthesis dataset}
    \label{tab:table1}
\end{table}

\clearpage

\subsection{Data Generation Pipeline}
To validate the diversity and applicability of the dataset, we conducted a statistical analysis of animation type distribution and temporal complexity. The dataset contains a total of 91,411 animation instances, with image-based animations accounting for 57.23\% and title and text animations making up 42.77\%. It covers 52 preset effects, including 42 entrance–exit pairs and 10 emphasis effects. The number of animation effects in the videos follows an approximately normal distribution, ranging from 4 to 15 effects, with videos containing 7 and 8 effects being the most frequent, each exceeding 5,000 instances. The durations of individual animation effects are most commonly 1.0s and 1.5s, accounting for over 55,000 instances.

\begin{figure}[h]
	\centering
	\includegraphics[width=0.7\textwidth]{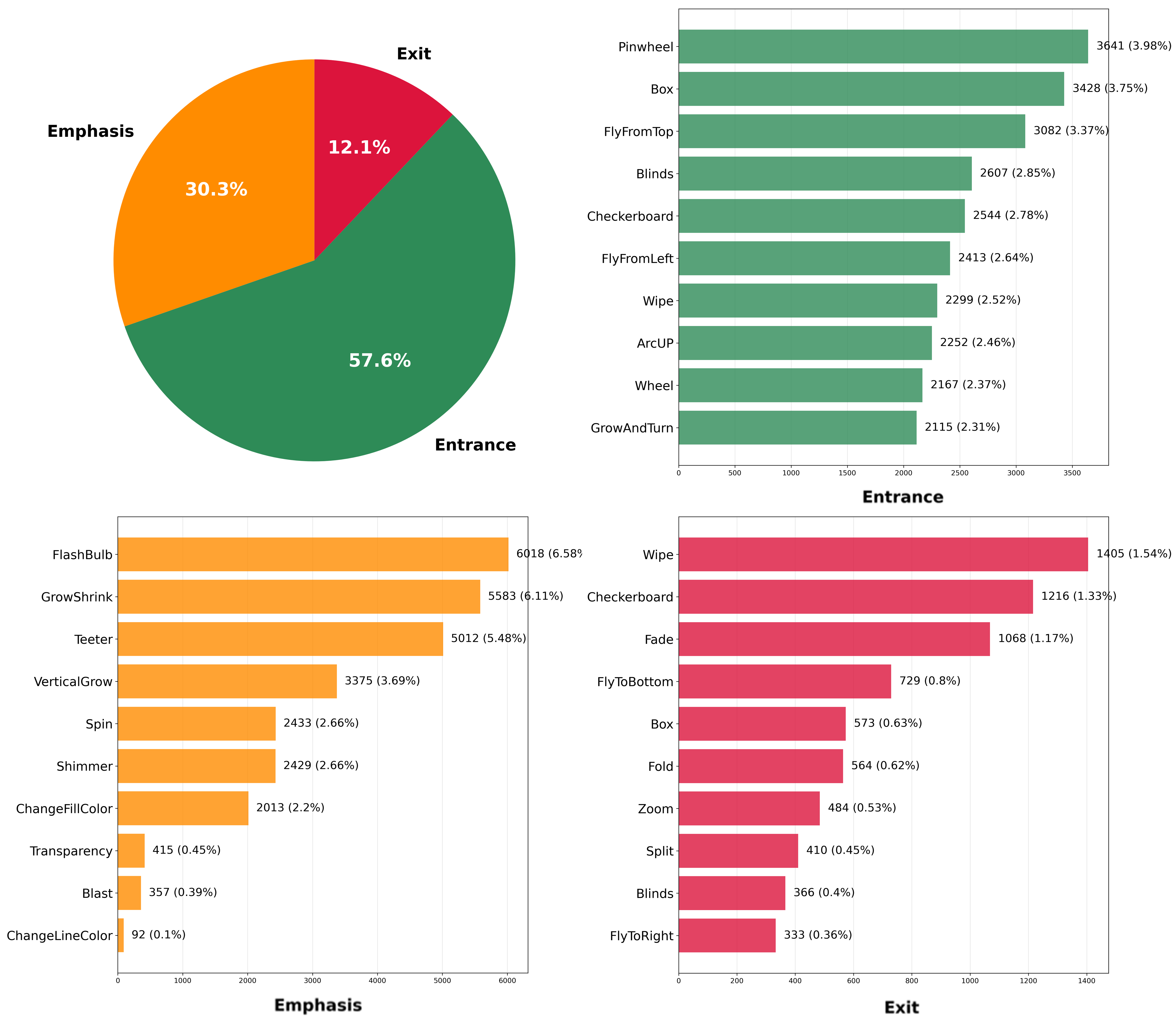}
	\caption{\textbf{Animation Type Distribution}: Pie chart and bar chart illustrating the overall proportions of entrance, emphasis, and exit animations. The bar chart also displays the top 10 animation effects (e.g., \textit{Pinwheel}, \textit{FlashBulb}, etc.) in terms of frequency and percentage for each category, with the y-axis representing animation types and the x-axis showing frequency/percentage.}
	\label{fig:fig3}
\end{figure}

LLMs exhibit distinct preferences for animation effects in text and images. For text entrances, the preferred effects are \textit{Box} (19.3\%) and \textit{Blinds} (15.2\%), which emphasize structured transitions. In contrast, image entrances favor effects like \textit{Pinwheel} (17.6\%) and \textit{FlyFromLeft} (11.9\%), which deliver a more dynamic impact. In terms of emphasis animations, text prefers \textit{Teeter} (33.5\%) and \textit{FlashBulb} (30.3\%), which highlight content, while images favor \textit{GrowShrink} (30.4\%) and \textit{Spin} (16.3\%), which focus on shape transformation. For exit animations, the top three effects are similar for both text and images: \textit{Wipe}, \textit{Checkerboard}, and \textit{Fade}, with text using these effects in the proportions of 22.4\%, 14.1\%, and 17.1\%, respectively, and images in 22.8\%, 21.2\%, and 17.3\%. 

Additionally, the LLM’s choice of animation details demonstrates consistency with common usage patterns (e.g., entering from above/left, exiting downward/right), such as a preference for \textit{FlyFromTop} in entrances (text: 15.8\%, image: 11.1\%) and \textit{FlyFromLeft} (image: 11.9\%) and a preference for \textit{FlyToBottom} (image: 12.5\%) and \textit{FlyToRight} (image: 6.5\%) in exits.

\begin{figure}[h]
	\centering
	\includegraphics[width=0.9\textwidth]{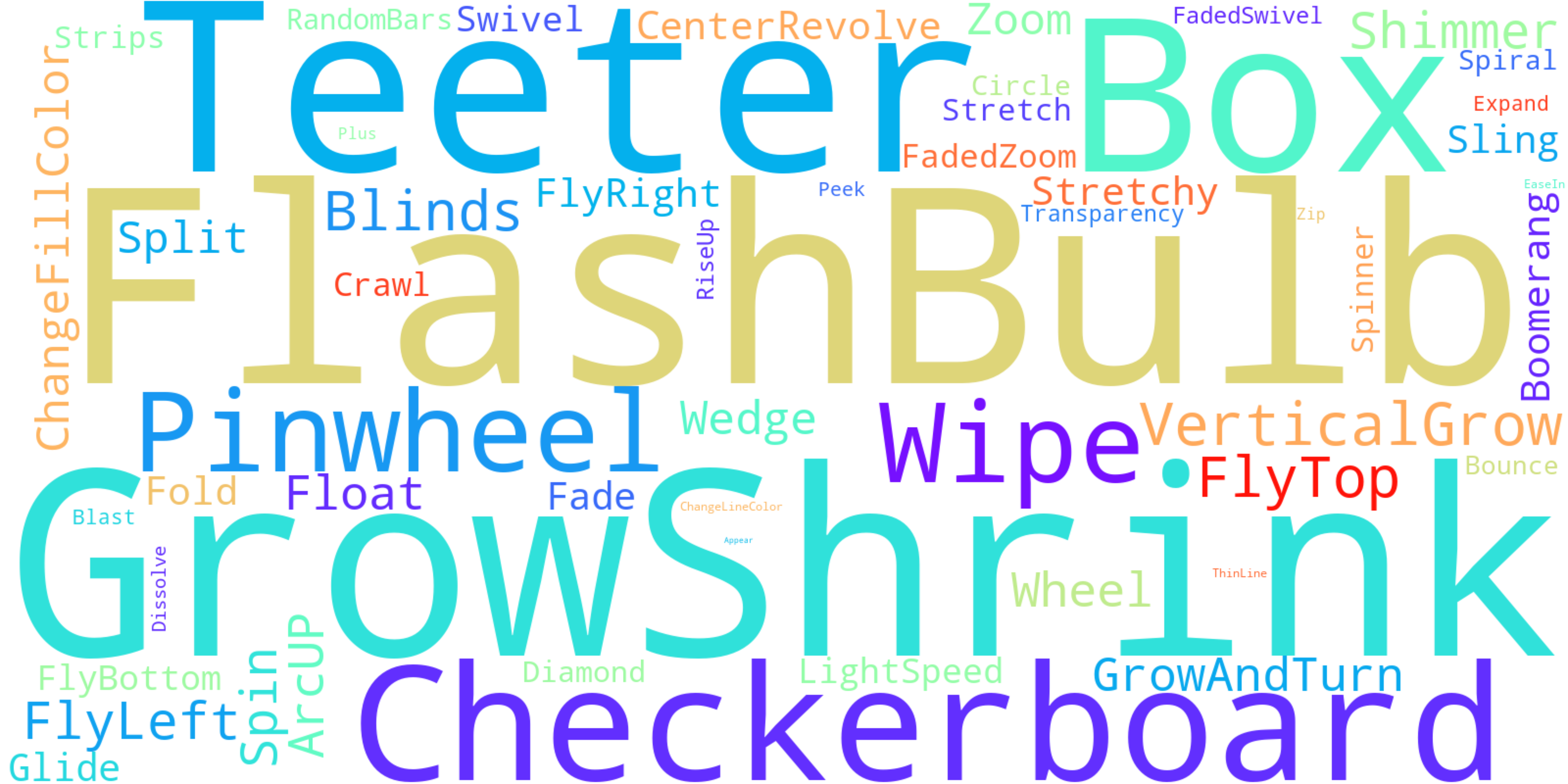}
	\caption{\textbf{Animation Type Word Cloud}: A word cloud displaying the frequency of all 52 animation types, with font size reflecting the frequency of occurrence. (Paired Entrance/Exit effects are merged.)}
	\label{fig:fig4}
\end{figure}

\begin{figure}[h]
	\centering
	\includegraphics[width=1.0\textwidth]{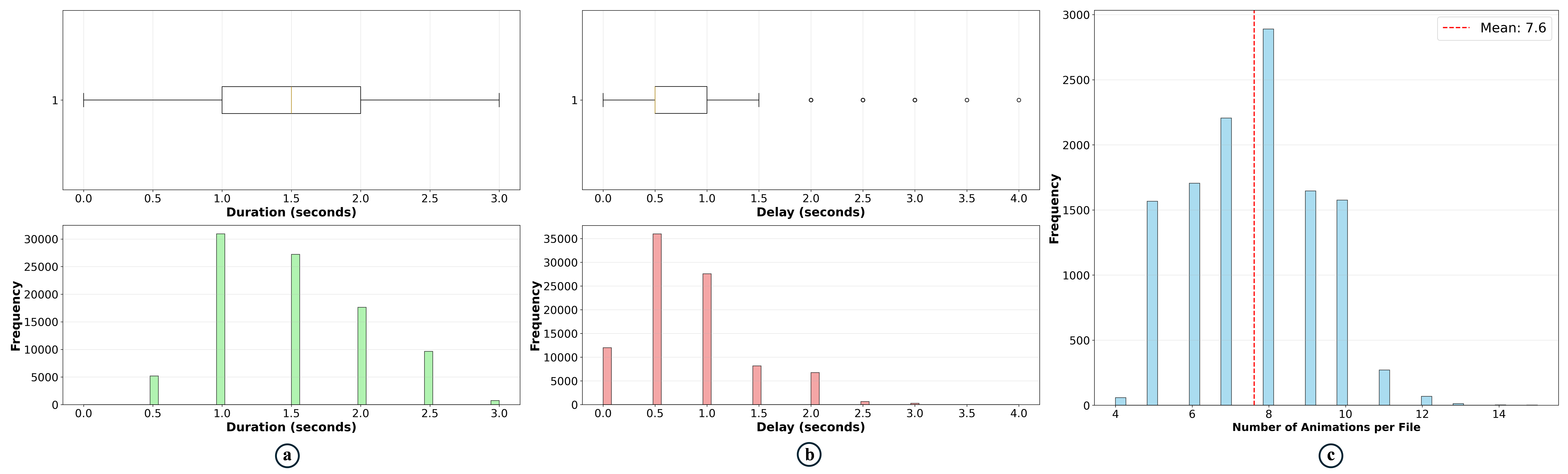}
	\caption{\textbf{(a) Duration Distribution Chart}: Displays the distribution of animation effect durations ranging from 0.5s to 3s. \textbf{(b) Delay Distribution Chart}: Shows the distribution of delays between animations, ranging from 0s to 4s. \textbf{(c) Number of Animations Distribution}: The number of animations per scheme follows an approximately normal distribution between 4 and 15, with an average of 7.6.}
	\label{fig:fig5}
\end{figure}

Overall, the animation types selected by the LLM are well-balanced, covering various scenarios such as entrances, emphasis, and exits, providing a diverse set of samples for model training. The analysis of temporal complexity shows that the animation durations and delays follow an approximately normal distribution, aiding the model in learning complex temporal dependencies.

\clearpage

\section{VLM Fine-Tuning}

The temporal dependencies and multimodal characteristics of slide animations pose significant challenges to visual language models (VLMs). To effectively recognize and describe animated slide transitions, we fine-tune Qwen 2.5-VL-7B using Low-Rank Adaptation (LoRA) on our synthetic dataset introduced in section 3. By aligning natural language animation descriptions with corresponding video sequences, the model’s understanding of temporal progression, motion types, and structural layout is substantially enhanced. This section details the rationale for model selection, the principles of LoRA-based fine-tuning, and the complete training pipeline, laying the methodological foundation for subsequent experiments.

\subsection{Model Selection}
The task of slide animation recognition demands that VLMs efficiently process video sequences and structured textual descriptions, requiring robust temporal modeling and multimodal fusion capabilities. Among various open-source VLMs, we select Qwen 2.5-VL for its architectural strengths, strong benchmarking performance, and active community support. Qwen 2.5-VL achieves competitive or superior results to closed-source models (e.g., GPT-4o, Gemini-1.5-Pro) on benchmarks such as Video-MME\cite{fu2025video}, MVBench\cite{li2024mvbench}, MMBench-Video\cite{fang2024mmbench}, and TempCompass\cite{liu2024tempcompass}, demonstrating its powerful temporal and multimodal reasoning abilities\cite{bai2025qwen2}.

Qwen 2.5-VL features a ViT-based visual encoder, which supports adaptive resolution input and dynamic frame rate sampling. This is particularly beneficial for processing animated slide videos with varying temporal structures. Moreover, its open-source ecosystem provides abundant resources—including model weights, documentation, and optimization tools—enabling rapid iteration and task-specific customization.

We choose the 7B parameter variant based on three considerations. First, it exhibits only marginal performance degradation compared to the 72B version, while outperforming other open-source models of similar scale such as LLaVA-7B and CLIP-ViT-L/14. Second, our hardware setup (4 × A800 80GB GPUs) supports efficient training at this scale, allowing for larger batch sizes that mitigate overfitting and significantly reduce training time. Third, our task emphasizes temporal modeling and action-level understanding rather than complex reasoning. The current limitations of VLMs on this task stem not from insufficient model capacity but from the lack of structured temporal animation data in pretraining. With LoRA-based fine-tuning, Qwen 2.5-VL-7B can effectively bridge this gap, improving its capacity to recognize motion types, temporal order, and structural elements in slide animations.

\subsection{LoRA}
Low-Rank Adaptation (LoRA)[8] is a parameter-efficient fine-tuning method that introduces trainable low-rank matrices into frozen pretrained weight matrices, enabling fast adaptation to downstream tasks with minimal computational and storage overhead. LoRA typically reduces trainable parameter count to 0.1\%–1\% of the full model, making it highly suitable for fine-tuning VLMs on specialized tasks such as slide animation recognition.

We adopt the LLaMA-Factory framework\cite{zheng2024llamafactory} to apply LoRA to the Transformer layers of Qwen 2.5-VL-7B, including self-attention and MLP modules. Pretrained weights remain frozen while LoRA layers are optimized, maximizing efficiency.

\subsubsection{Mathematical Formulation}
The core idea of LoRA is that weight updates in a pretrained model can be well-approximated by a low-rank formulation, thereby significantly reducing the parameter space required for optimization. While traditional full fine-tuning directly updates the weight matrix $W$, LoRA approximates the weight change using a low-rank decomposition, keeping the original weight $W_0$  frozen and introducing a trainable low-rank update $\Delta W$. Mathematically, the weight update in LoRA is represented as:

\begin{equation}
W = W_0 + \Delta W
\end{equation}

This update is further decomposed into two low-rank matrices as $\Delta W=A\cdot B$, where $A$ and $B$ denote the low-rank matrices. The dimensions of $A \in \mathbb{R}^{d \times r}$  and  $B \in \mathbb{R}^{r \times k}$ are governed by the rank $r$, which is much smaller than $min(d,k)$ . Both $A$ and $B$ are initialized with Gaussian distributions at the beginning of training. Given an input $x$, the model output becomes:

\begin{equation}
y = W x = (W_0 + \Delta W) x = W_0 x + (A \cdot B) x
\end{equation}

During training, the actual update follows the rule: $\Delta W = \frac{\alpha}{r} A \cdot B$, where the scaling factor \textit{lora\_alpha} (denoted as $\alpha$) regulates the update magnitude to prevent overfitting. This decomposition reduces the number of trainable parameters from $d\cdot k$ to $(d+k)\cdot r$ , thus significantly lowering the optimization cost of fine-tuning Qwen 2.5-VL-7B while maintaining its adaptability to the downstream task.

\subsubsection{Implementation Details}
In Qwen 2.5-VL-7B, LoRA is applied to target modules such as the self-attention components (\texttt{q\_proj}, \texttt{k\_proj}, \texttt{v\_proj}) and the MLP components (\texttt{up\_proj}, \texttt{down\_proj}, \texttt{gate\_proj}), covering the main weight matrices in the 28-layer model structure. LoRA configurations include \texttt{rank}=8/16, \texttt{lora\_alpha}=16, and \texttt{lora\_dropout}=0.05. Training hyperparameters involve the AdamW optimizer (learning rate of 5e-5 with cosine scheduling), a per-GPU batch size of 4 (at 4 FPS) or 8 (at 1 or 2 FPS), and 5 epochs of training.

For slide animation videos with a sampling rate of 2 FPS, a single 80GB GPU supports training on 8 video samples per batch. Gradient accumulation is performed every 2 steps to optimize resource usage, allowing one full fine-tuning cycle to be completed in approximately 17 hours—significantly lower than the time required for full-parameter fine-tuning. The modular design also facilitates transfer learning, enabling future expansion to larger models or more animation types.

\subsection{Fine-Tuning Pipeline}
We leverage Qwen 2.5-VL’s built-in support for adaptive resolution and dynamic frame rates\cite{bai2025qwen2}, thus requiring no additional preprocessing for multimodal inputs. The supervised training dataset consists of 11,000 paired samples of natural language descriptions and corresponding animation videos (as detailed in Section 3). To effectively capture the temporal features of animations and the semantic relationships between textual descriptions, we employ Cross-Entropy Loss as the objective function to optimize the model's performance in generation tasks, due to its proven effectiveness in measuring token-level prediction accuracy.

This study also investigates the effects of various configurations, including three frame rates (1, 2, and 4 FPS) and two LoRA ranks (8 and 16). Training was conducted over the course of one week, with continuous real-time monitoring using SwanLab, which tracked loss curves, learning rate schedules, and GPU memory usage. Detailed visual representations of the training metrics are provided in \textbf{Appendix: Training Records}.

\clearpage

\section{Metric Design and Experiments}

This section systematically validates the effectiveness of the LoRA fine-tuning approach in the task of slide animation recognition through experimental design and data analysis. The experiments cover both the synthetic dataset detailed in \textbf{Section 3} and a manually created slide animation test set. Through comparative analysis, ablation studies, and multi-dimensional evaluation, this section aims to verify the feasibility of the fine-tuning method, analyze the performance differences of general pre-trained models in this specialized domain, and compare the performance of the fine-tuned smaller specialized model with that of leading closed-source models. Additionally, the impact of various hyperparameter configurations on fine-tuning performance is explored. Furthermore, this section assesses the generalization capability of the fine-tuned model on the manually created test set, providing solid empirical evidence for future model optimization.

\subsection{Experimental Setup}
To comprehensively evaluate the effectiveness of the LoRA-based fine-tuning strategy in the Qwen 2.5-VL-7B slide animation recognition task, we designed a systematic experimental process. First, in terms of dataset construction, we partitioned 1,000 paired samples of natural language descriptions and corresponding animation videos described in Section 3, which were not used in training, as a test set for independent evaluation using automated metrics and CODA scoring. Additionally, we manually created and annotated 50 slides animation videos to assess the model's generalization capability in heterogeneous scenarios.

In terms of model configuration, we use the original Qwen 2.5-VL-7B as the baseline and fine-tune it with different frame rates (1, 2, and 4 FPS) and LoRA rank combinations (8 and 16). This setup aims to explore the interaction between frame rate and rank on recognition performance. Additionally, we independently trained a Rank 32 model (at 4 FPS) to evaluate the potential performance improvement from a higher rank.

To compare the performance differences between open-source and industry-leading closed-source models, we selected GPT-4.1 and Gemini-2.5-Pro as benchmarks to assess the relative advantages and limitations of various models in the slide animation description generation task.

Regarding evaluation metrics, we consider both quantitative assessments of generated descriptions in terms of linguistic form and semantic accuracy, as well as qualitative measures of content completeness and coherence. The automated evaluation uses classic metrics such as BLEU\cite{papineni2002bleu}, ROUGE\cite{lin2004rouge}, SPICE\cite{anderson2016spice} to quantify the match between model outputs and descriptions at both the literal and semantic levels. Additionally, we leverage our custom-built CODA scoring framework to evaluate the results along three dimensions: action coverage, temporal order, and detail fidelity. The specific scoring method for CODA will be detailed in Section 5.2.

\subsection{Metric Design}
This section provides a detailed definition of the automated metrics and the CODA large model scoring method, as well as their roles in the slide animation description generation task, ensuring the scientific rigor and reproducibility of performance evaluation.

The automated metrics include BLEU-4\cite{papineni2002bleu}, ROUGE\cite{lin2004rouge}, SPICE\cite{anderson2016spice}, which quantify the quality of the generated descriptions across three dimensions: language fluency, content coverage, and semantic similarity.

BLEU-4 quantifies the fluency of the generated description by calculating the overlap of 4-grams between the generated and reference descriptions. Its formula is as follows:
\begin{equation}
    \text{BLEU-4} = \text{BP} \cdot \exp\left(\sum_{n=1}^{4} w_n \log p_n\right)\
\end{equation}
where $p_n$ represents the precision for $n$-grams ($n$ from 1 to 4) in the generated description, which is calculated as the number of matching n-grams divided by the total number of $n$-grams in the generated description, with the maximum count for each n-gram adjusted by the reference descriptions.$w_n$ is the weight for each n-gram precision, typically set as a uniform distribution ($w_n=\frac{1}{4}$).$BP$ is the brevity penalty factor, which is defined as:
\begin{equation}
    \text{BP} = 
        \begin{cases} 
        1 & \text{if } c > r, \\
        e^{(1 - r/c)} & \text{if } c \leq r
        \end{cases}
\end{equation}
In this context, $c$ represents the length of the generated description, and $r$ represents the length of the reference description that is closest to $c$.

The ROUGE series of metrics are based on recall and assess content coverage through word-level (ROUGE-1), bigram-level (ROUGE-2), and longest common subsequence (ROUGE-L) overlap. The formulas for these metrics are as follows:
\begin{equation}
    \text{ROUGE-N} = \frac{\sum_{S \in \{\text{Reference Sentences}\}} \sum_{\text{gram}_n \in S} \text{Count}_{\text{match}}(\text{gram}_n)}{\sum_{S \in \{\text{Reference Sentences}\}} \sum_{\text{gram}_n \in S} \text{Count}(\text{gram}_n)}
\end{equation}
where $Count_{match}(gram_n)$ represents the count of matching $n$-grams, and $Count(gram_n)$ is the total count of $n$-grams in the reference description. ROUGE-L is specifically defined as: $\text{ROUGE-L} = \frac{\text{LCS}(R, G)}{\text{length}(R)}$,  where $LCS(R,G)$ is the length of the longest common subsequence between the reference description $R$ and the generated description $G$.

SPICE is based on semantic graph matching and evaluates the semantic similarity between the generated and reference descriptions. The formula is: $SPICE=F_1(TP,FP,FN)$ ,where $F_1$ is the harmonic mean of precision and recall:
\begin{equation}
    F_1=2\cdot\frac{Precision\cdot Recall}{Precision + Recall}    
\end{equation}
$TP$, $FP$, and $FN$ represent the true positives, false positives, and false negatives in semantic graph matching, respectively.

CODA large model scoring evaluates the generated descriptions qualitatively, focusing on three dimensions: action coverage, temporal order, and detail fidelity. The scoring process includes decomposing both the predicted description (Prediction) and the reference description (Reference) into ordered action unit sequences $P=[p_1,...,p_m]$ and $R=[r_1,...,r_n]$, where an action unit is the smallest semantic segment (e.g., "Image A fades in"). A left-to-right nearest-match strategy is used to form a matching pair set $M$. The metric calculation is as follows:

\textbf{Coverage}: $Coverage=\frac{|M|}{n}$ , measures the proportion of matched action units relative to the reference description.

\textbf{Order}: Based on the length of the \textit{Longest Common Subsequence (LCS)}, $L$:
\begin{equation}
    Order = \frac{L}{n} (\text{if }n=0, \text{ set Order}=1)    
\end{equation}
which reflects the temporal accuracy.

\textbf{Detail}: For each $(r_i ,p_j )\in M$ , compare action parameters (such as count, direction, etc.). A perfect match scores 1, a partial match scores 0.5, and a complete mismatch scores 0. Detail score is the average value (if no match is found, the score is 0).

\subsection{Comparative Analysis}
\textbf{LoRA-tuned Qwen 2.5-VL-7B achieves the highest scores on every automated and CODA metric, establishing a clear state-of-the-art for synthetic slide-animation caption.}

This section compares the LoRA model with baseline and closed-source systems on a 1 000-video synthetic test set, evaluated by BLEU-4, ROUGE-L, SPICE, and CODA (Coverage, Order, Detail). To suppress irrelevant generations, we apply task-specific prompts (\textbf{Appendix: Prompts}) and fix the LLM temperature at 0.0; each CODA score is averaged over three repetitions for robustness.

The original Qwen 2.5-VL-7B performs poorly across frame rates—only marginally surpassing GPT-4.1 at 4 FPS—and fails to capture temporal and structural animation cues. In contrast, LoRA fine-tuning markedly improves language fluency, content coverage, and semantic alignment, with the rank-32/4 FPS setting yielding the largest gains in action coverage, temporal order,and detail fidelity.

After fine-tuning, LoRA outperforms closed-source models. While Gemini-2.5-Pro slightly exceeds GPT-4.1 at higher frame rates ($\approx $ 10 \% CODA lead), both trail the LoRA model on every metric. Notably, GPT-4.1 scores better at 2 FPS than at 4 FPS, whereas LoRA maintains consistent improvements across frame rates and rank values.

\begin{figure}
	\centering
	\includegraphics[width=0.9\textwidth]{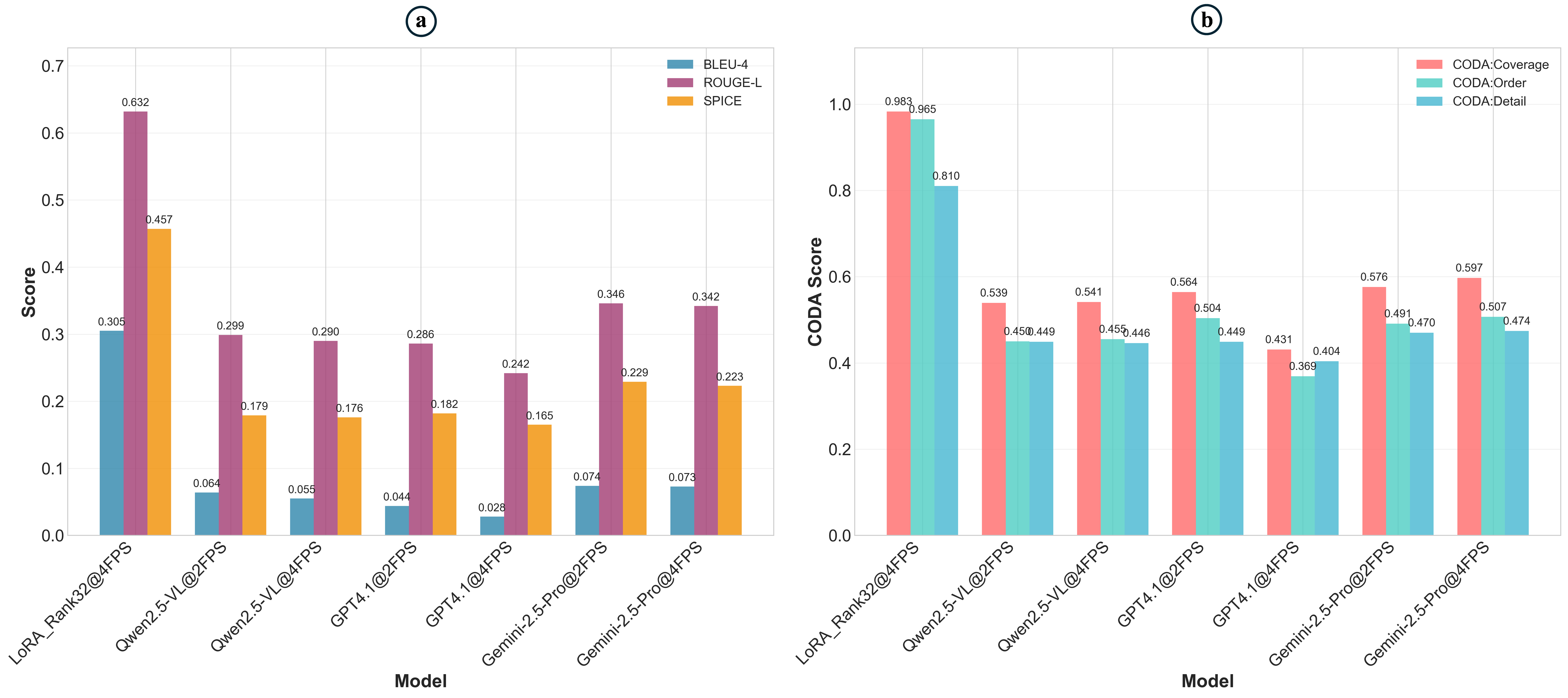}
	\caption{Performance of different models on the 1,000-video synthetic test set. The leftmost bars correspond to our model. \textbf{(a) Automated metrics} (BLEU-4, ROUGE-L, SPICE); \textbf{(b) CODA metrics} (Coverage, Order, Detail).}
	\label{fig:fig6}
\end{figure}

\textbf{Figure \ref{fig:fig6}a} contrasts automated metrics, \textbf{Figure \ref{fig:fig6}b} shows CODA results, and \textbf{Table \ref{tab:table2}} lists detailed scores, collectively confirming the superiority of LoRA fine-tuning and motivating subsequent ablation and generalization studies.

\begin{table}[ht]
    \caption{Performance comparison of different models on the synthetic test set}
    \centering
    \begin{tabular}{@{}lccccccc@{}}
        \toprule
        Model Configuration & BLEU-4$\uparrow$ & ROGUE-L$\uparrow$ & SPICE$\uparrow$ & CODA:Coverage$\uparrow$ & CODA:Order$\uparrow$ & CODA:Detail$\uparrow$ \\ \midrule
        \textbf{LoRA\_Rank32@4FPS}  & \textbf{0.305} & \textbf{0.632} & \textbf{0.457} & \textbf{0.983} & \textbf{0.965} & \textbf{0.810} \\
        Qwen2.5-VL-7B@2FPS         & 0.064 & 0.299 & 0.179 & 0.539 & 0.450 & 0.449 \\
        Qwen2.5-VL-7B@4FPS         & 0.055 & 0.290 & 0.176 & 0.541 & 0.455 & 0.446 \\
        GPT4.1@2FPS                & 0.044 & 0.286 & 0.182 & 0.564 & 0.504 & 0.449 \\
        GPT4.1@4FPS                & 0.028 & 0.242 & 0.165 & 0.431 & 0.369 & 0.404 \\
        Gemini-2.5-Pro@2FPS        & 0.074 & 0.346 & 0.229 & 0.576 & 0.491 & 0.470 \\
        Gemini-2.5-Pro@4FPS        & 0.073 & 0.342 & 0.223 & 0.597 & 0.507 & 0.474 \\
        \bottomrule
    \end{tabular}
    \label{tab:table2}
\end{table}

\clearpage

\subsection{Ablation Study}
This ablation adopts the same test bed as Section 5.3, varying frame rate (FPS $\in$ {1, 2, 4}) and LoRA rank (8, 16, 32) to isolate their effects on slide-animation recognition.

\textbf{Higher frame rates drive the largest performance gains}. Performance rises monotonically from 1 FPS to 4 FPS. At 4 FPS the model attains its best BLEU-4 and ROUGE-L scores and peaks on all three CODA facets, confirming that denser temporal sampling captures animation dynamics more completely.

\textbf{Increasing LoRA rank yields only marginal returns beyond rank 16.} At any given FPS the gap between ranks 8, 16, and 32 is modest. Rank 32 improves detail fidelity slightly at 4 FPS, but at 1 FPS rank 8 outperforms rank 16, suggesting that high-rank adapters may overfit when temporal context is sparse. Rank-selection should therefore track the chosen frame rate for an optimal cost–performance balance.

\begin{figure}[ht]
	\centering
	\includegraphics[width=0.9\textwidth]{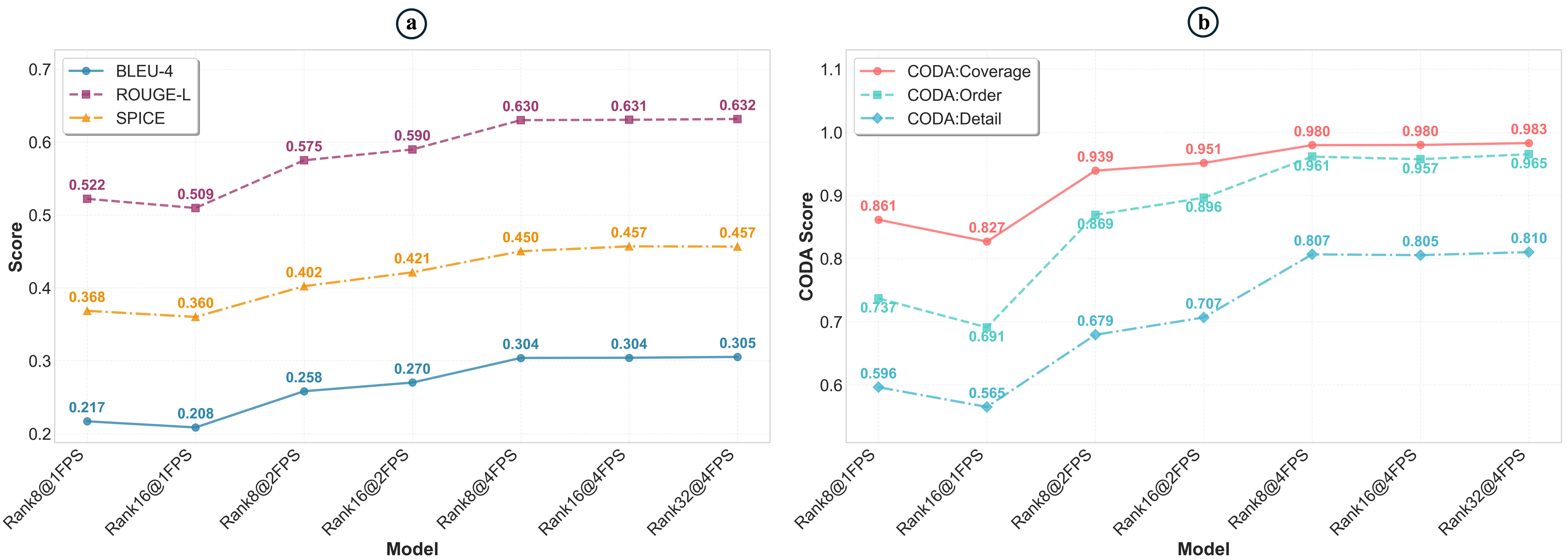}
	\caption{Performance variations on the synthetic test set across different frame rates and LoRA ranks, from Rank 8 at 1 FPS up to Rank 32 at 4 FPS.  \textbf{(a) Automated metrics} (BLEU-4, ROUGE-L, SPICE); \textbf{(b) CODA metrics} (Coverage, Order, Detail).}
	\label{fig:fig7}
\end{figure}

\textbf{Figure \ref{fig:fig7}a} plots automated metrics across FPS settings, while \textbf{Figure \ref{fig:fig7}b} traces CODA scores across ranks; \textbf{Table \ref{tab:table3}} lists the full results for all seven configurations, providing a quantitative basis for these observations.

\begin{table}[ht]
    \caption{Performance Comparison of Different LoRA Configurations on the Synthetic Dataset}
    \centering
    \begin{tabular}{@{}lccccccc@{}}
        \toprule
        Model Configuration & BLEU-4$\uparrow$ & ROGUE-L$\uparrow$ & SPICE$\uparrow$ & CODA:Coverage$\uparrow$ & CODA:Order$\uparrow$ & CODA:Detail$\uparrow$ \\ \midrule
        LoRA\_Rank8@1FPS   & 0.217 & 0.522 & 0.368 & 0.861 & 0.737 & 0.596 \\
        LoRA\_Rank16@1FPS  & 0.208 & 0.509 & 0.360 & 0.827 & 0.691 & 0.565 \\
        LoRA\_Rank8@2FPS   & 0.258 & 0.575 & 0.402 & 0.939 & 0.869 & 0.679 \\
        LoRA\_Rank16@2FPS  & 0.270 & 0.590 & 0.421 & 0.951 & 0.896 & 0.707 \\
        LoRA\_Rank8@4FPS   & 0.304 & 0.630 & 0.450 & 0.980 & 0.961 & 0.807 \\
        LoRA\_Rank16@4FPS  & 0.304 & 0.631 & 0.457 & 0.980 & 0.957 & 0.805 \\
        \textbf{LoRA\_Rank32@4FPS}  & \textbf{0.305} & \textbf{0.632} & \textbf{0.457} & \textbf{0.983} & \textbf{0.965} & \textbf{0.810} \\
        \bottomrule
    \end{tabular}
    \label{tab:table3}
\end{table}

\subsection{Generalization Performance Analysis}
\textbf{LoRA fine-tuning still achieves the best performance on a 50-video manually created test set.} However, its performance margin narrows due to the presence of previously unseen animation patterns in the new data, which diminishes the benefits of higher frame rates.

The generalization benchmark contains 50 video–description pairs manually created by multiple volunteers. We evaluate LoRA-Rank 16 @ 2 FPS and LoRA-Rank 32 @ 4 FPS against the original Qwen 2.5-VL-7B and the closed-source GPT-4.1 and Gemini-2.5-Pro (each at 2 FPS and 4 FPS) with BLEU-4, ROUGE-L, SPICE, and CODA (Coverage, Order, Detail).

\textbf{LoRA remains on top}: Both LoRA variants substantially exceed the baseline on every metric: for example, LoRA-Rank 32 raises CODA Coverage from 0.404 to 0.574 and CODA-Order from 0.352 to 0.477. These gains confirm that low-rank adaptation also improves in manually created test set.

\textbf{Lead reduced by data mismatch}: Compared with results on the synthetic set (Section 5.3), LoRA's improvements shrink because the manually created slides contain background motions, concurrent effects, and layouts that the training data never exhibited. Such distribution shifts also damp the frame-rate benefit: moving from 2 FPS to 4 FPS adds only 0.02–0.03 BLEU points here, versus $\geq$ 0.05 on synthetic data.

\begin{figure}
	\centering
	\includegraphics[width=0.9\textwidth]{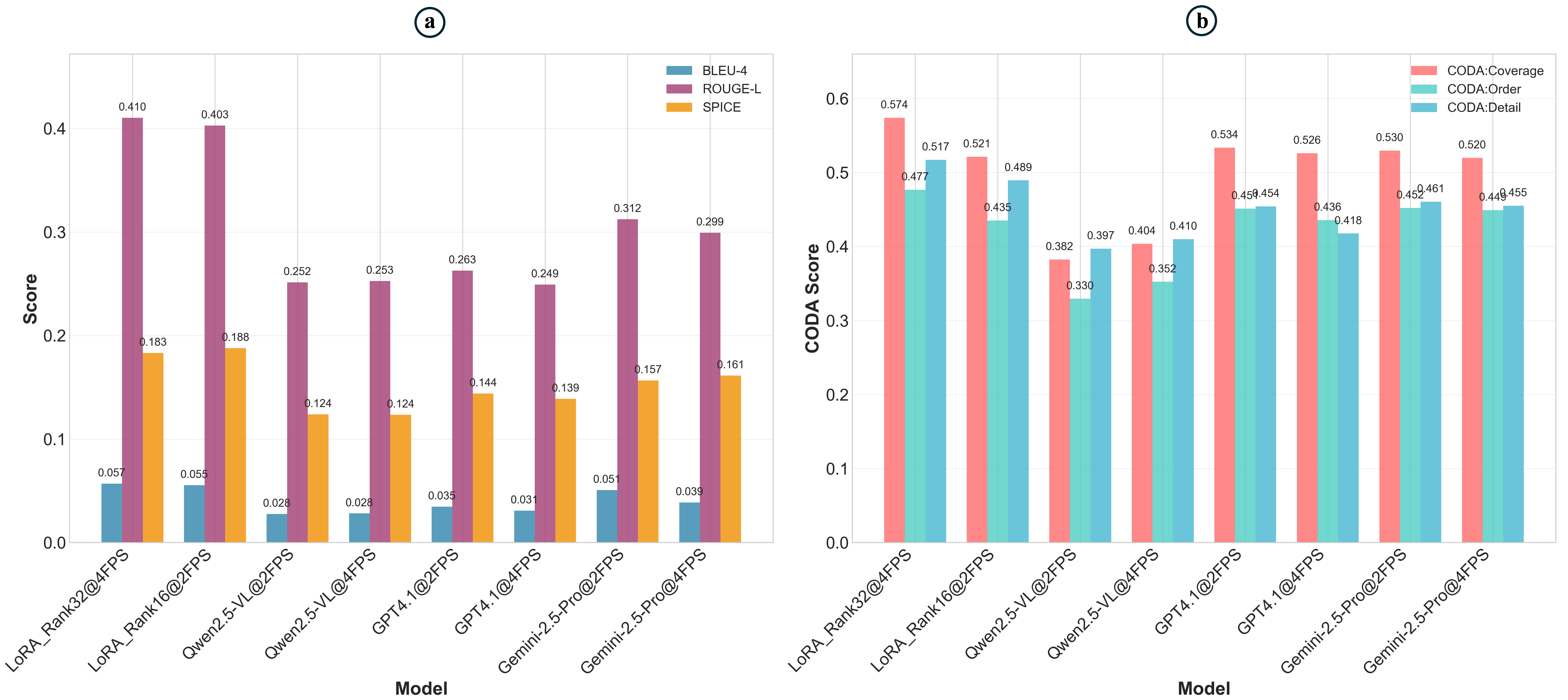}
	\caption{Performance of different models on the 50-video manually created test set. Left two bars correspond to our model. \textbf{(a) Automated metrics} (BLEU-4, ROUGE-L, SPICE); \textbf{(b) CODA metrics} (Coverage, Order, Detail).}
	\label{fig:fig8}
\end{figure}

\textbf{Figures \ref{fig:fig8}a–b} plot automated and CODA metrics, and \textbf{Table \ref{tab:table4}} lists all scores, underscoring that while LoRA generalizes best to manually created slides, further diversity in training data and refined adaptation strategies will be required to widen the gap under real-world variability.

\begin{table}[ht]
    \caption{Performance comparison of different models on the manually created test set}
    \centering
    \begin{tabular}{@{}lccccccc@{}}
        \toprule
        Model Configuration & BLEU-4$\uparrow$ & ROGUE-L$\uparrow$ & SPICE$\uparrow$ & CODA:Coverage$\uparrow$ & CODA:Order$\uparrow$ & CODA:Detail$\uparrow$ \\ \midrule
        GPT4.1@2FPS        & 0.035 & 0.263 & 0.144 & 0.534 & 0.451 & 0.454 \\
        GPT4.1@4FPS        & 0.031 & 0.249 & 0.139 & 0.526 & 0.436 & 0.418 \\
        Gemini-2.5-Pro@2FPS & 0.051 & 0.312 & 0.157 & 0.530 & 0.452 & 0.461 \\
        Gemini-2.5-Pro@4FPS & 0.039 & 0.299 & 0.161 & 0.520 & 0.449 & 0.455 \\
        Qwen2.5-VL@2FPS    & 0.028 & 0.252 & 0.124 & 0.382 & 0.330 & 0.397 \\
        Qwen2.5-VL@4FPS    & 0.028 & 0.253 & 0.124 & 0.404 & 0.352 & 0.410 \\
        \textbf{LoRA\_Rank16@2FPS}  & 0.055 & 0.403 & \textbf{0.188} & 0.521 & 0.435 & 0.489 \\
        \textbf{LoRA\_Rank32@4FPS}  & \textbf{0.057} & \textbf{0.410} & 0.183 & \textbf{0.574} & \textbf{0.477} & \textbf{0.517} \\
        \bottomrule
    \end{tabular}
    \label{tab:table4}
\end{table}

\clearpage

\section{Conclusion and Future Work}

\subsection{Conclusion}
This work conducts a comprehensive investigation of slide animation generation and proposes a systematic solution, covering the pipeline, data processing, and model training aspects. First, we release the first public slide-animation dataset—12,000 text–JSON–video triplets—built with an end-to-end synthesis framework that directly tackles the field’s data-scarcity problem. Second, we fine-tune Qwen 2.5-VL-7B using our synthesized dataset, demonstrating substantial improvements in recognition accuracy. Finally, we introduce \textbf{Coverage–Order–Detail Assessment (CODA)}, a novel evaluation metric specifically designed for slide animation recognition tasks. Extensive experiments demonstrate the effectiveness of fine-tuning on our synthetic dataset, with the resulting models exhibiting generalization performance on manually created test set. 

\subsection{Limitations \& Future Work}
Despite these advancements, several limitations remain. First, the insufficient semantic richness of static slides continues to pose a fundamental constraint on model performance. Second, constrained by limited computational resources, our exploration of frame rate (FPS) and rank-related parameters remains inadequate, leaving considerable room for further improvement. In future work, we aim to enhance the synthetic pipeline with more sophisticated page composition logic, enriching the dataset with greater content variability and structural complexity. We will further investigate temporal modeling in visual encoders—exploring the effectiveness of ViT frame sampling for animation dynamics—and explore policy-guided fine-tuning with Group Relative Policy Optimization (GRPO)\cite{shao2024deepseekmath} to boost adaptation efficiency and generalization.

\section*{Author Contributions}
\textbf{Yifan Jiang} led the synthetic pipeline design and implement, fine-tuning experiments, and manuscript writing (except related work). \textbf{Yibo Xue} conducted early motion effect experiments, literature review, wrote the related work section, and created data charts. \textbf{Yukun Kang} assisted with experimental code and flowchart design. \textbf{Pin Zheng} prepared the manually created slide test set and contributed to experimental design and data analysis. \textbf{Jian Peng} provided overall guidance and manuscript revision. \textbf{Feiran Wu} offered expert advice on algorithmic optimization and evaluation strategies, strengthening the rigor and robustness of the study. \textbf{Changliang Xu} identified key innovative directions and offered methodical guidance throughout the research process. All authors have read and approved the final manuscript.

\bibliographystyle{IEEEtran}
\bibliography{IEEEabrv,references}

\end{document}